# INRIASAC: Simple Hypernym Extraction Methods

Gregory Grefenstette
Inria
1 rue Honoré d'Estienne d'Orves
91120 Palaiseau, France
`Gregory.grefenstette@inria.fr`

## Abstract

For information retrieval, it is useful to classify documents using a hierarchy of terms from a domain. One problem is that, for many domains, hierarchies of terms are not available. The task 17 of SemEval 2015 addresses the problem of structuring a set of terms from a given domain into a taxonomy without manual intervention. Here we present some simple taxonomy structuring techniques, such as term overlap and document and sentence co-occurrence in large quantities of text (English Wikipedia) to produce hypernym pairs for the eight domain lists supplied by the task organizers. Our submission ranked first in this 2015 benchmark, which suggests that overly complicated methods might need to be adapted to individual domains. We describe our generic techniques and present an initial evaluation of results.

## 1 Introduction

This paper describes two simple hypernym extraction methods, given a list of domain terms and a large amount of text divided into documents. Task 17 of the 2015 Semeval campaign (Bordea *et al*., 2015) consists in structuring a flat list of pre-identified domain terms into a list of hypernym pairs. Task organizers provide two lists of terms for each of four domains: *equipment, food, chemical, science*, one extracted from WordNet and one from an unknown source. Participants in the task were allowed to use any resource (except existing taxonomies) to automatically transform the lists of terms into lists of pairs of terms, the first term being a hyponym of the more general second term. For example, if the words `airship` and `blimp` were included in the lists of terms for a domain, the system was expected to return lines such as:

```
       blimp         airship
```

The task organizers provided training data from the domains of Artificial Intelligence, vehicles and plants, different from the test domains. The training data consisted in term lists (for plants), and term lists and lists of hypernyms (for AI and for vehicles). We examined these files to get an understanding of the task but did not exploit them.

We used the English text of Wikipedia (downloaded from http://dumps.wikimedia.org on August 13, 2014) as our only resource for discovering these relations. We extracted only the text of each article, ignoring titles, section headings, categories, infoboxes, or other meta-information present in the article. We recognized task terms in these articles and gathered statistics on document and sentence co-occurrence between domain terms, as well as term frequency. To recognize hypernyms, we used term inclusion (explained in section 3.1 below) and co-occurrence statistics (see section 3.2) to decide whether two terms were possibly in a hypernym relation, and document frequency to chose which term was the hypernym. Our submission ranked first in the SemEval 2015 task 17 benchmark.

## 2 Domain Lists

Participants were provided with the eight lists of domain terms, each containing between 370 and 1555 terms. Some terms examples:

**chemical**: agarose, nickel sulfate heptahydrate, aminoglycan, pinoquercetin, …
**equipment:** storage equipment, strapping, traveling microscope, minneapolis-moline, …
**food:** sauce gribiche, botifarra, phitti, food colouring, bean, limequat, kalach, …
**science:** biological and physical, history of religions of eastern origins, linguistic anthropology, religion, semantics…
**WN_chemical**: abo antibodies, acaricide, acaroid resin, acceptor, acetal, acetaldehyde…
**WN_equipment**: acoustic modem, aerator, air search radar, amplifier, anti submarine rocket, apishamore, apparatus, …
**WN_food**: absinth, acidophilus milk, adobo, agar, aioli, alcohol, ale, alfalfa, allemande, allergy diet, …
**WN_science**: abnormal psychology, acoustics, aerology, aeromechanics, aeronautics, …

Terms consisted of one to nine words. Some terms were very short and ambiguous (only two or three characters: `ga, os, tu, ada, aji, …`) and some very long (e.g., `udp-n-acetyl-alpha-d-muramoyl-l-alanyl-gamma-d-glutamyl-l-lysyl-d-alanyl-d-alanine, korea advanced institute of science and technology satellite 4`). It is specified that the taxonomies produced during the task should be rooted on `chemical` for the two chemical domain lists, on `equipment` for the equipment lists, on `food` for the food lists, and on `science` for the science lists, even though the term `chemical` was absent from the WN_chemical domain list. Participants were allowed to add additional nodes, i.e. terms, in the hierarchy as they consider appropriate. We did not add any new terms, except for `chemical` in the WN_chemical list.

## 2.1 Preprocessing the resource

Our only resource for discovering hypernym relations was the English Wikipedia. Starting from the wiki-latest-pages-articles.xml, we extracted all the text between <text> markers, and marked off document boundaries using <title> markers. No other information (infoboxes, categories, etc.) was kept. The text was then tokenized and output as one sentence per line. The first English Wikipedia sentence extracted looked like this: `' Anarchism ' is a political philosophy that advocates stateless societies often defined as self-governed voluntary institutions , but that several authors have defined as more specific institutions based on non-hierarchical free associations.` We applied Porter stemming (Willet, 2006) and replaced stopwords (Buckley *et al.*, 1995) by underscores. The first sentence then becomes:
```
anarch _ _ _ polit philosophi _ advoc
stateless societi _ defin _ self-govern
voluntari institut _ _ _ sever author _
defin _ _ specif institut base _ non-
hierarch free associ
```

We applied the same Porter stemming and stopword removal to the task-supplied domain terms. So the *science* term list, for example, becomes

```
0 electro-mechan system
1 biolog _ physic
2 histori _ religion _ eastern origin
3 linguist anthropolog
4 metaphys
```

We retained both Porter-stemmed versions of the Wikipedia sentences and domain terms as well as the original unstemmed versions for the treatment described below.

## 3 Extracting Hypernyms

In order to extract hypernyms, we used the following features: (i) presence of terms in the same sentence, (ii) presence in the same document (iii) term frequency (iv) document frequency, and (v) subsequences.

### 3.1 Subterms

In addition to domain lists supplied for the Semeval task, we were supplied with training data. One file in this training data, ontolearn_AX.taxo, gives ground truth for the training file ontolearn_AX.terms, and contains:

```
source code    < code
theory of inheritance < theory
```

From these validated examples, we concluded that an 'easy' way to find hypernyms is to check whether one term is a suffix of the other (e.g., `communications satellite` as a type of `satellite`), or whether one term B is the prefix of another term B A C where A is any two-letter word (e.g. `helmet of coţofeneşti` as a type of `helmet`; `caterpillar d9` as a type of `caterpillar`). We chose two letters for the second term to cover English prepositions such as *of, in, by*, … This heuristic was unexpectedly productive in the chemical domain where many hypernym pairs were similar to: `ginsenoside mc` as a type of `ginsenoside` (see Table 1). But our prefix matching using second words of length two missed hypernyms such as `fortimicin b` as a type of `fortimicin` or `ginsenoside c-y` as a type of `ginsenoside`. Obviously chemical terms should have their own heuristics for subterm matching.

Other examples of errors, false positives, caused by these heuristics are `licorice` as a type of `rice` or `surface to air missile system` as a type of `surface`.

## 3.2 Sentence and Document Co-occurrence Statistics

For other domain terms (which could include the hypernyms found by the suffix and prefix heuristics), we use the statistics of document presence, and of co-occurrence of terms in sentences to predict hypernym relations. Let $D_{porter}(term)$ be the document frequency of a Porter-stemmed term in the stemmed version of Wikipedia. Since Wikipedia article boundaries were stored, we considered each Wikipedia article as a new document. Let $SentCooc_{porter}(term_i, term_j)$ be the number of times that the Porter-stemmed versions of $term_i$ and $term_j$ appear in the same sentence in the stemmed English Wikipedia. Given two terms, $term_i$ and $term_j$, if $term_i$ is appears in more documents than $term_j$, then $term_i$ is a candidate hypernym for $term_j$.

CandHypenym($term_i$) = { $term_j$ :
$SentCooc_{porter}(term_i, term_j) > 0$ &&
$D_{porter}(term_j) > D_{porter}(term_i)$   }

This heuristically derived set is meant to capture the intuition that general terms are more widely distributed than more specific terms (e.g., `dog` appears in more Wikipedia articles than `poodle`).

| Domain | suffix | prefix | cooc | Total hypernyms produced |
|---|---|---|---|---|
| WN_chemical | 750 | 10 | 3766 | 4001 |
| WN_equipment | 171 | 3 | 1338 | 1369 |
| WN_food | 616 | 25 | 4121 | 4238 |
| WN_science | 174 | 0 | 1070 | 1102 |
| chemical | 10780 | 91 | 19322 | 28443 |
| equipment | 241 | 17 | 1126 | 1168 |
| food | 471 | 33 | 4277 | 4363 |
| science | 193 | 17 | 1130 | 1164 |

Table 1. Number of prefix and suffix hypernyms produced, compared to the total number of hypernyms returned for each domain.

Next, we define the best hyperym candidate for $term_i$ as being the term $term_k$ that appears in the most documents (from Wikipedia in this case):

BestHypernym($term_i$) = $term_k$
such that
$\forall\ term_j \in$ CandHypernym($term_i$) :
$D_{porter}(term_k) \geq D_{porter}(term_j)$

Next, we remove this term $term_k$ from CandHypernym($term_i$) and repeat the heuristic twice, retaining, then, the three candidate hypernyms appearing in the most documents for each term not found by using the prefix or suffix heuristics.

### 3.2.1 Co-occurrence Example

Consider the following example. In the domain file science.terms there is the term `biblical studies`. The Porter-stemmed version of this term `biblic studi` appears in 887 documents. Considering all the other terms in science.terms, we find that `biblic studi` appears 215 times in the same sentence as the stemmed version of theology (`theologi`), 111 times in the same sentences as stemmed history (`histori`), 50 times with religion, 43 times with music, and 42 times with science (`scienc`).

```
215 887 21977     biblic studi    theologi
111 887 383927    biblic studi    histori
50  887 64044     biblic studi    religion
43  887 412791    biblic studi    music
```

We decided to keep the top three for simplicity, so this term contributed three bolded lines above to our submitted science.taxo file.

## 3.3 Other Attempts at Finding Relations

We tried a number of other methods to find hypernyms, none of which gave results that looked good from a cursory glance. We implemented a method to recognize sentences containing Hearst patterns (list from (Cimiano *et al.*, 2005)) involving the domain terms. For example, `tape` is in *equipment*, and were able to find stemmed sentences of the form A, B and other C … such as `todai , sticki note , 3m #tape# @, and other@ #tape# ar exampl of psa ( pressure-sensit adhes )` from which we should have been able to extract relations such as `3m tape` is a type of `tape`, and `sticky note` is a type of `tape`. But we would have had to the parse the sentence, and been willing to add new terms (which was permitted by the organizers) to the derived hypernym lists but we did not want to make that processing investment yet. We also tried to discover the *basic vocabulary* (Kit, 2002) of each domain without success.

## 4 Evaluation

Each participant in Task 17 of SemEval 2015 was allowed to submit one run for each of the 8 domains (see Table 1 for the names of the domains, and the number of hypernym pairs we submitted. Suffix and prefix subterms account for 10% to 36% of the hypernyms we produced. The cooccurrence technique produced the most hypernym candidates). The task organizers evaluated the submissions of the six participating teams, using automated and manual methods, and published their evaluation three weeks after the submission deadline. Our team placed first in the official ranking of the six teams.

| Domain | suffix | prefix | cooc | union | gold to find |
|---|---|---|---|---|---|
| WN_chemical | 377 | 5 | 574 | 644 | 1387 |
| WN_equipment | 119 | 0 | 168 | 184 | 485 |
| WN_food | 371 | 2 | 681 | 726 | 1533 |
| WN_science | 119 | 0 | 230 | 240 | 441 |
| chemical | 2019 | 9 | 715 | 2407 | 24817 |
| equipment | 184 | 1 | 286 | 305 | 615 |
| food | 279 | 1 | 807 | 822 | 1587 |
| science | 121 | 7 | 193 | 209 | 465 |

Table 2. Number of gold standard relations to find in the last column. Columns 2, 3 and 4 are the number of gold standard relations found by each technique. "union" is the union of columns 2, 3 and 4. Since the co-occurrence technique can find relations that have been found by the suffix and prefix techniques.

| Domain | suffix | prefix | cooc | union | gold to find |
|---|---|---|---|---|---|
| WN_chemical | 26% | 0.3% | 40% | 46% | 1387 |
| WN_equipment | 24% | 0% | 34% | 38% | 485 |
| WN_food | 23% | 0.1% | 43% | 47% | 1533 |
| WN_science | 26% | 0% | 51% | 54% | 441 |
| chemical | 8% | 0.03% | 3% | 10% | 24817 |
| equipment | 30% | 0.02% | 47% | 50% | 615 |
| food | 18% | 0.06% | 51% | 52% | 1587 |
| science | 26% | 1.8% | 42% | 45% | 465 |

Table 3. Percentage of correct answers found by each method.

The evaluation criteria, which were not published before the submission, combined the presences of cycles in the hypernyms submitted, the Fowlkes & Mallows measure of the overlap between the submitted hierarchy and the gold standard hierarchy, the F-score ranking, the number of domains submitted (not all teams returned results for all domains), and a manual precision ranking (for hypernyms not present in the gold standard). The gold standards used by the task organizers came from published taxonomies, or from subtrees of WordNet (prefixed as WN_ above). A quick evaluation of how well our simple hypernym extraction techniques fared on each gold standard is shown in Table 2.

As Table 3 shows, most of the correct answers found come from the sentence and document co-occurrence method described in section 3.2.

## 5 Conclusion

Even though training data was provided for this taxonomy creation task, we did not exploit it in this our first participation in Semeval. We implemented some simple frequency-based co-occurrence statistics, and substring inclusion heuristics to propose a set of hypernyms. We did not implement any graph algorithms (cycle detection, branch deletion) that would be useful to build a true hierarchy. Future plans involve examining and eliminating cycles generated by this method. Since we only used wikipedia as a resource, the method depends on the given terms being present in Wikipedia, which was not always the case, especially in the chemical domain. In future work, we will also examine using web documents, in lieu of or to supplement Wikipedia.

### Acknowledgments

This research is partially funded by a research grant from INRIA, and the Paris-Saclay Institut de la Société Numérique funded by the IDEX Paris-Saclay, ANR-11-IDEX-0003-02.